\documentclass[10pt,twocolumn,letterpaper]{article}

\usepackage{cvpr}              

\usepackage{graphicx}
\usepackage{grffile}
\usepackage{amsmath}
\usepackage{amssymb}
\usepackage{booktabs}
\usepackage{multirow}
\usepackage{xcolor}
\usepackage{colortbl} 
\def\figvspace{{\vspace{-3mm}}}

\usepackage[export]{adjustbox}

\usepackage{mathtools}

\usepackage[pagebackref,breaklinks,colorlinks,citecolor=blue,linkcolor=blue,urlcolor=blue]{hyperref}

\usepackage[capitalize]{cleveref}
\crefname{section}{Sec.}{Secs.}
\Crefname{section}{Section}{Sections}
\Crefname{table}{Table}{Tables}
\crefname{table}{Tab.}{Tabs.}

\def\cvprPaperID{1431} 
\def\confName{CVPR}
\def\confYear{2022}

\begin{document}
	
	\title{Video Frame Interpolation Transformer}
	
	\author{
		Zhihao Shi$^{*1}$
		\and
		Xiangyu Xu$^{*\dagger2}$
		\and
		Xiaohong Liu$^{3}$
		\and
		Jun Chen$^{1}$
		\and
		Ming-Hsuan Yang$^{4,5,6}$
		\\
		\and		
		$^1$McMaster Univeristy
		\and
		$^2$Nanyang Technological University
		\and
		$^3$Shanghai Jiao Tong University
		\and
		$^4$University of California, Merced
		\and
		$^5$Yonsei University
		\and
		$^6$Google Research
}


\maketitle

\def\thefootnote{*}
\footnotetext{These authors contributed equally.}
\def\thefootnote{$\dagger$}
\footnotetext{Corresponding author.}
\def\thefootnote{\arabic{footnote}}

\begin{abstract}
	Existing methods for video interpolation heavily rely on deep convolution neural networks, and thus suffer from their intrinsic limitations, such as content-agnostic kernel weights and restricted receptive field. 
	To address these issues, we propose a Transformer-based video interpolation framework that allows content-aware aggregation weights and considers long-range dependencies with the self-attention operations.
	To avoid the high computational cost of global self-attention, we introduce the concept of local attention into video interpolation and extend it to the spatial-temporal domain. 
	Furthermore, we propose a space-time separation strategy to save memory usage, which also improves performance.
	In addition, we develop a multi-scale frame synthesis scheme to fully realize the potential of Transformers.
	Extensive experiments demonstrate the proposed model performs favorably against the state-of-the-art methods both quantitatively and qualitatively on a variety of benchmark datasets.
	The code and models are released at \url{https://github.com/zhshi0816/Video-Frame-Interpolation-Transformer}.	
\end{abstract}

\section{Introduction}
\label{sec:intro}
Video frame interpolation aims to temporally upsample an input video by synthesizing new frames between existing ones.
It is a fundamental problem in computer vision that involves the understanding of motions, structures, and natural image distributions, which facilitates numerous downstream applications, such as image restoration~\cite{brooks2019learning, xu2017motion}, virtual reality~\cite{anderson2016jump}, and medical imaging~\cite{karargyris2010three}.


Most of the state-of-the-art video frame interpolation methods are based on deep convolution neural networks (CNNs)~\cite{niklaus2017video, niklaus2018video, lee2020adacof, shi2021video,jiang2018super, park2020bmbc, bao2019depth, xu2019quadratic}.
While achieving the state-of-the-art performance, these CNN-based architectures usually suffer from two major drawbacks. 
First, the convolution layer is content-agnostic, where the same kernels are used to convolve with different locations of different inputs.
While this design can serve as a desirable inductive bias for image recognition models to acquire translational equivalence~\cite{krizhevsky2012imagenet}, 
it is not always suitable for video interpolation which involves a complex motion-compensation process that is spatially-variant and content-dependent.
Thus, adopting CNN backbones may restrict the ability of adaptive motion modeling and potentially limits further development of video interpolation models.

\begin{figure}[t]
	\centering
	\begin{minipage}[h]{\linewidth}
		\centering
		\includegraphics[width=.98\linewidth]{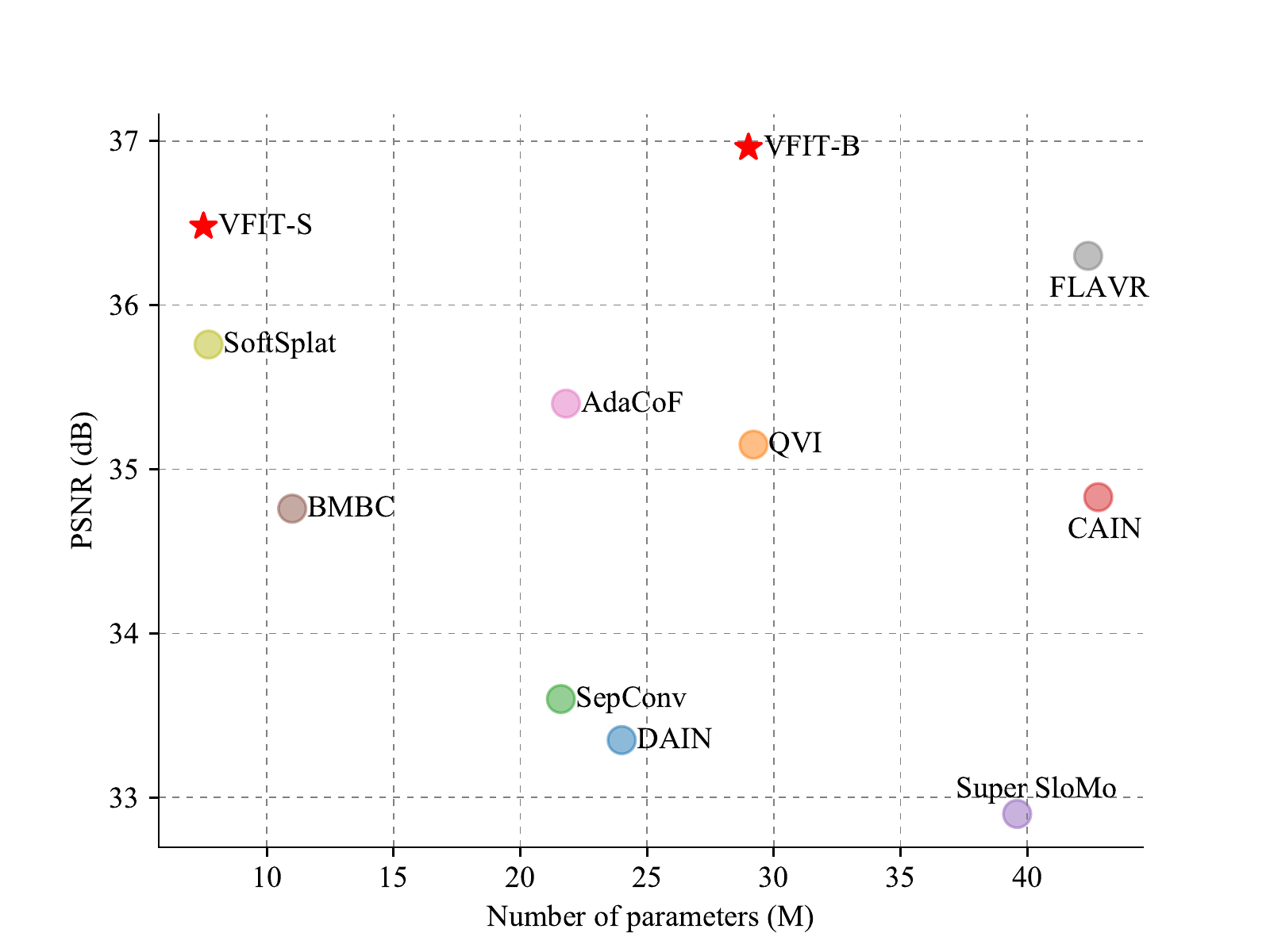}
	\end{minipage}
	\figvspace
	\caption{{Comparison of performance and model size} using the Vimeo-90K dataset~\cite{xue2019video}. 
		VFIT outperforms the state-of-the-art methods with fewer parameters. 
		VFIT-S and VFIT-B denote the proposed small and base models.
	}
	\vspace{-5mm}
	\label{fig:performance}
\end{figure}

Second, capturing long-range dependencies is of central importance in video interpolation for which large motion fields pose the most prominent challenges. 
However, most CNNs~\cite{xu2019quadratic,lee2020adacof} usually employ small convolution kernels (typically 3$\times$3 as suggested by VGG~\cite{simonyan2014very}), which is inefficient in exploiting long-range information and thus less effective in synthesizing high-quality video frames.
While it seems an easy fix to use larger kernels in the convolution layer, it significantly increases the number of model parameters and computational cost, thereby leading 
to poor local minimums in training without proper regularizations.
Moreover, simply stacking multiple small kernel layers for a larger receptive field does not fully resolve this problem either, as distant dependencies cannot be effectively
learned in a multi-hop fashion~\cite{wang2018non}.


On the other hand, Transformers~\cite{vaswani2017attention}, which are initially designed for natural language processing (NLP) to efficiently model long-range dependencies between input and output, naturally overcome the above drawbacks of CNN-based algorithms, and are in particular suitable for the task of video interpolation.
Motivated by the success in NLP, several methods recently adapt Transformers to computer vision and demonstrate promising results on various tasks, such as image classification~\cite{dosovitskiy2020image, touvron2021training}, semantic segmentation~\cite{wang2020axial}, object detection~\cite{carion2020end}, and 3D reconstruction~\cite{xu2021texformer}.
Nevertheless, how to effectively apply Transformers to video interpolation that involves an extra temporal dimension remains an open yet challenging problem. 


In this work, we propose the Video Frame Interpolation Transformer (VFIT) for effective video interpolation.
Compared with typical Transformers~\cite{carion2020end,dosovitskiy2020image,chen2021pre} where the basic modules are largely borrowed from the original NLP model~\cite{vaswani2017attention}, there are three distinguished designs in the proposed VFIT to generate photo-realistic and temporally-coherent frames.
First, the original Transformer~\cite{vaswani2017attention} is based on a self-attention layer that interacts with the input elements (e.g., pixels) globally.
As this global operation has quadratic complexity with regard to the number of elements, directly applying it to our task leads to extremely high memory and computational cost due to the high-dimensionality nature of videos.
Several methods~\cite{chen2021pre,cao2021video} circumvent this problem by dividing the feature maps into patches and treating each patch as a new element in the self-attention. 
However, this strategy cannot model fine-grained dependencies between pixels inside each patch which are important for synthesizing realistic details.
Moreover, it may introduce edge artifacts around patch borders.
In contrast, we introduce the local attention mechanism of Swin~\cite{liu2021swin} into VFIT to address the complexity issue while retaining the capability of modeling long-range dependencies with its shift-window scheme.
We demonstrate that with proper development and adaptation, the local attention mechanism originally used for image recognition can effectively improve the video interpolation performance with a smaller amount of parameters as shown in Figure~\ref{fig:performance}.

Second, the original local attention mechanism~\cite{liu2021swin} is only suitable for image input and cannot be easily used for the video interpolation task where an extra temporal dimension is involved. 
To address this issue, we generalize the concept of local attention to spatial-temporal domain, 
which leads to the Spatial-Temporal Swin attention layer (STS) that is compatible with videos.
However, this simple extension could lead to memory issues when using large window sizes.
To make our model more memory efficient, we further devise a space-time separable version of STS, called Sep-STS, by factorizing the spatial-temporal self-attention.
Interestingly, Sep-STS not only effectively reduces memory usage but also considerably improves video interpolation performance.

To exploit the full potential of our Sep-STS, we propose a new multi-scale kernel-prediction framework which can better handle multi-scale motion and structures in diverse videos, and generates high-quality video interpolation results in a coarse-to-fine manner.
%
%
%
%
%
%
%
%
The proposed VFIT is concise, flexible, light-weight, high-performing, fast, and memory-efficient.
As shown in Figure~\ref{fig:performance}, a small model (VFIT-S) already outperforms the state-of-the-art FLAVR method~\cite{kalluri2020flavr} by 0.18 dB with only \textbf{17.7\%} of its parameters, while our base model (VFIT-B) achieves \textbf{0.66 dB} improvement with 68.4\% of its parameters.

	
	
	

\section{Related Work}
%

\noindent\textbf{Video frame interpolation.}
Existing video frame interpolation methods can be broadly classified into three categories: flow-based~\cite{jiang2018super, park2020bmbc, bao2019depth, xu2019quadratic,sim2021xvfi}, kernel-based~\cite{niklaus2017video, niklaus2018video, lee2020adacof,niklaus2021revisiting}, and direct-regression-based methods~\cite{karargyris2010three}.


The flow-based methods~\cite{jiang2018super, park2020bmbc, bao2019depth, xu2019quadratic} generate intermediate frames by warping pixels from the source images 
according to predicted optical flow. 
Although these methods perform well, 
they are usually based on simplified motion assumptions such as linear~\cite{jiang2018super} and quadratic~\cite{xu2019quadratic},
limiting their performance in many real-world scenarios where the assumptions are violated.


Unlike the flow-based approaches, the kernel-based methods \cite{niklaus2017video, niklaus2018video, lee2020adacof,niklaus2021revisiting} do not rely on any prescribed assumptions and thus generalize better to diverse videos. 
For example, SepConv~\cite{niklaus2018video} predicts adaptive separable kernels to aggregate source pixels of the input, and AdaCoF~\cite{lee2020adacof} learns deformable spatially-variant kernels that are used to convolve with the input frames to produce the target frame.
However, these approaches usually apply the kernel prediction modules at one scale and thereby cannot effectively handle complex motions and structures that could appear in different scales.
Moreover, these CNN-based methods do not account for long-range dependency among pixels.
In contrast, we propose a multi-scale Transformer-based kernel prediction module, which achieves higher-quality results for video interpolation as will be shown in Section~\ref{sec:exp}.

\begin{figure*}[t]
	\centering
	\begin{minipage}[h]{\linewidth}
		\centering
		\includegraphics[width=1\linewidth]{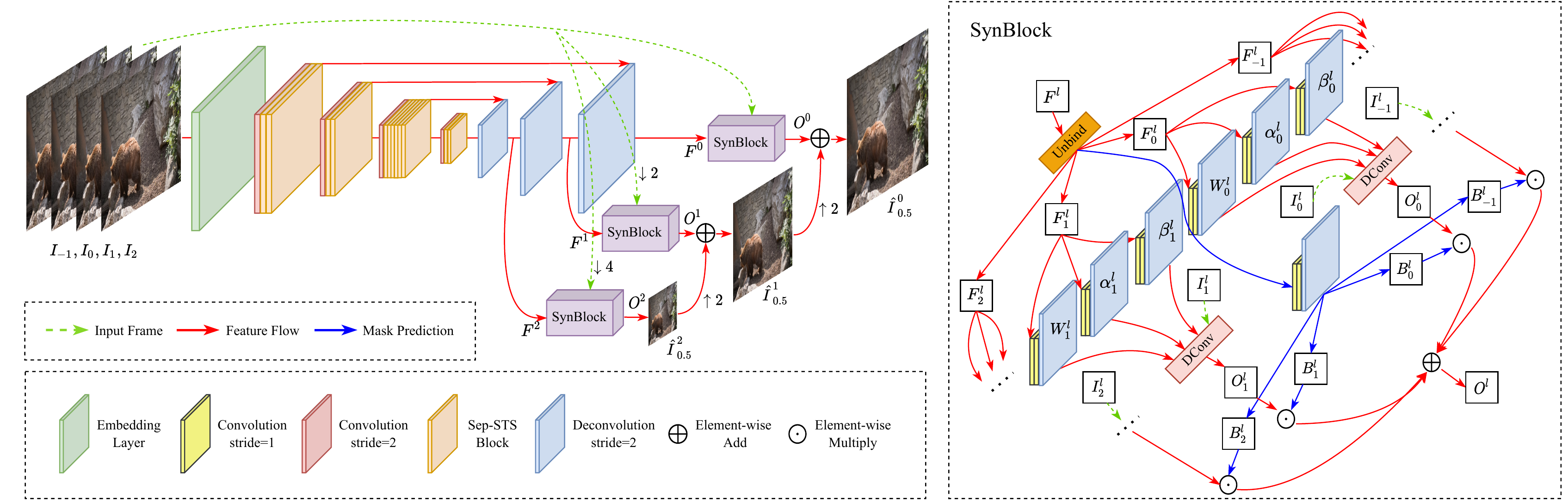}
	\end{minipage}
	\figvspace
	\caption{{Overview of the proposed VFIT.} We first use an embedding layer to transform input frames into shallow features, followed by a Transformer-based encoder-decoder network to extract deep hierarchical features. 
		These features together with the input frames are fed to a multi-scale frame synthesis network that is composed of three SynBlocks to obtain the final output.
		``$\downarrow n$'' and ``$\uparrow n$'' denote downsampling and upsampling by a factor of $n$, respectively.
		``DConv'' represents the generalized deformable convolution in~\cite{stpan}.
		Note the SynBlock can be seen as a multi-frame extension of AdaCoF~\cite{lee2020adacof} originating from STPAN~\cite{stpan}.
		Please find more detailed explanations in Section~\ref{sec:method}.
	}
	\vspace{-3mm}
	\label{fig:FrameWork}
\end{figure*}

Recently, Kalluri~\etal~\cite{kalluri2020flavr} propose a CNN model to directly regress the target frame, which achieves the state-of-the-art results. 
As shown in Figure~\ref{fig:performance}, the proposed VFIT outperforms this method by a clear margin with fewer parameters, which clearly shows the advantages of Transformers in video interpolation.


\vspace{1mm}
\noindent\textbf{Vision Transformer.}
Transformers have recently been applied to various vision tasks, such as visual classification~\cite{dosovitskiy2020image,liu2021swin,tran2018closer, xie2018rethinking, gberta_2021_ICML}, object detection~\cite{carion2020end}, semantic segmentation~\cite{wang2020axial}, 3D reconstruction~\cite{xu2021texformer}, and image restoration~\cite{chen2021pre}. 
%
Nevertheless, it has not been exploited for video frame interpolation. 
%
In this work, we propose VFIT that achieves the state-of-the-art performance with a light-weight model.
To overcome the high computational cost caused by global self-attention, we introduce the local attention mechanism of Swin~\cite{liu2021swin} to avoid the  complexity issues while retaining the ability of long-range dependency modeling.
We note that one concurrent work~\cite{liang2021swinir} also uses the local attention for low-level vision tasks.
However, it only considers image input and cannot deal with videos which are more challenging to handle due to the extra temporal dimension.
In contrast, we extend the concept of local attention to the spatial-temporal domain to enable Transformer-based video interpolation, and propose a space-time separation strategy which not only saves memory usage but also acts as an effective regularization for performance gains.

\section{Proposed Method}
\label{sec:method}
Figure~\ref{fig:FrameWork} shows an overview of the proposed model.
Similar to existing methods~\cite{niklaus2017video,niklaus2018video,xu2019quadratic,kalluri2020flavr}, to synthesize an intermediate frame $I_{0.5}$,
we use its $T$ neighboring frames $I_{\{-(\lfloor{\frac{T}{2}}\rfloor-1), \cdots, 0, 1, \cdots, {\lceil{\frac{T}{2}}\rceil}\}}$ as the input.
Specifically, the input frames are $I_{-1},I_0,I_1,I_2$ when $T$ is $4$.
%

The proposed VFIT consists of three modules: shallow feature embedding, deep feature learning, and final frame synthesis.
First, the embedding layer takes the input frames and generates shallow features for  the deep feature learning module.
Similar to \cite{liu2021swin}, the shallow embedding is realized with a convolution layer,
where we adopt the 3D convolution rather than its 2D counterpart in \cite{liu2021swin} to better encode the spatial-temporal features of the input sequence.
Next, we feed the shallow features to the deep module to extract hierarchical feature representations $\{F^l, l=0,1,2\}$ to capture the multi-scale motion information (Section~\ref{sec:deep feature learning}).
Finally, an intermediate frame $\hat{I}_{0.5}$ can be generated by the frame synthesis blocks (SynBlocks in Figure~\ref{fig:FrameWork}) using the deep features $F^l$ (Section~\ref{sec:frame synthesis}).

\begin{figure*}[t]
	\hspace{2mm}
	\centering
	\begin{minipage}[h]{0.95\linewidth}
		
		\centering
		\includegraphics[width=\linewidth]{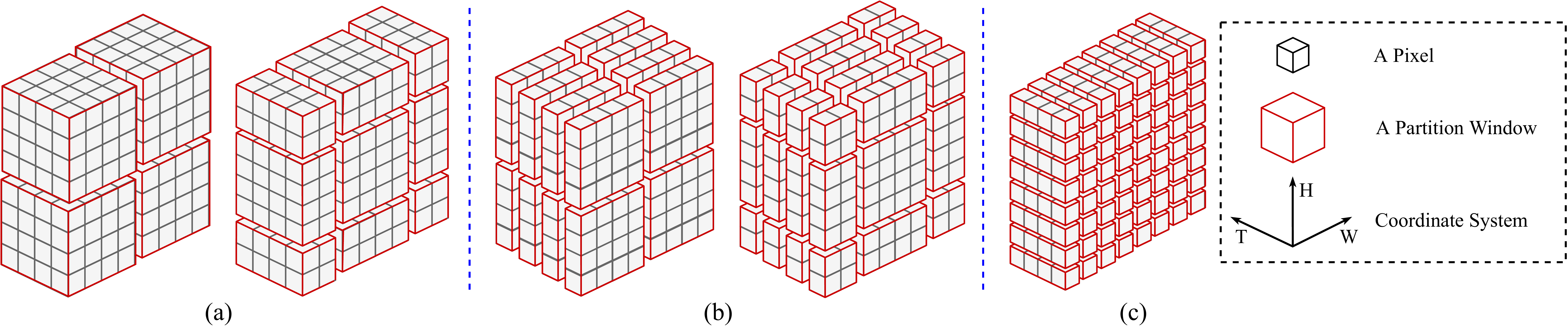}
	\end{minipage}
	\figvspace
	\caption{{Illustration of different local partition strategies.}
		(a) Regular and shifted partitions for spatial-temporal cubes of STS. 
		(b) Regular and shifted partitions for spatial windows of Sep-STS. 
		(c) Temporal vector partition for Sep-STS. 
	}
	\vspace{-3mm}
	\label{fig:partition}
\end{figure*}

\begin{figure}[t]
	\hspace{2mm}
	\centering
	\begin{minipage}[h]{0.98\linewidth}
		\centering
		\includegraphics[width=\linewidth]{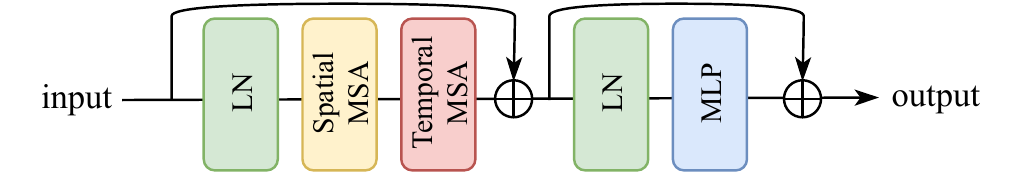}
	\end{minipage}
	\vspace{-2mm}
	\caption{{Illustration of the Sep-STS block.} The Spatial MSA and Temporal MSA represent the multi-head self-attentions in spatial and temporal local windows, respectively (Section~\ref{sec:deep feature learning}).
	}
	\vspace{-3mm}
	\label{fig:blocks arch}
\end{figure}

\subsection{Learning Deep Features}  \label{sec:deep feature learning}
As shown in Figure~\ref{fig:FrameWork}, we use a Transformer-based encoder-decoder architecture for learning features.  
The encoder is composed of four stages, where each stage starts with a 3D convolution layer using a stride of 2 to downsample the input features, and the downsampling layer is followed by several Sep-STS blocks which are the main components of our framework. 
For the decoder, we use a light-weight structure that only has three 3D deconvolution layers with a stride of 2 to upsample the low-resolution feature maps. 
Note that we only resize the spatial dimension of the features throughout our network and leave the temporal size unchanged.
Next, we provide more explanations about the proposed Sep-STS block.


\vspace{1mm}
\noindent \textbf{Local attention.}
Existing Transformers~\cite{vaswani2017attention,dosovitskiy2020image,carion2020end} mainly adopt a global attention mechanism to aggregate information from the input, which could cause extremely high memory and computational cost for video frame interpolation.
A straightforward solution to this problem is to directly divide the feature maps into patches and treat each patch as a new element in the global attention~\cite{cao2021video,chen2021pre}.
This strategy is equivalent to aggressively downsampling the input with pixel shuffle~\cite{shi2016real} (downsampling factor equals to the patch size) and cannot well reconstruct high-quality image details which require fine-grained dependency modeling between pixels. 

In this work, we introduce the local attention mechanism of Swin Transformer~\cite{liu2021swin}, which can effectively address the above issues.
First, as the self-attention of Swin is computed inside local windows, it naturally avoids the heavy computational burden of global attentions.
Second, Swin employs a shifted-window partition strategy to connect different local regions, and alternatively using regular and shifted-window partitions enables long-range dependency modeling.
Nevertheless, this method is designed for image applications and cannot be easily applied to videos.

\vspace{1mm}
\noindent \textbf{Spatial-temporal local attention.} 
To make the Swin Transformer compatible with video inputs, we generalize the local attention mechanism to spatial-temporal space and propose the STS attention. 
As shown in Figure~\ref{fig:partition}(a),
the STS is conceptually similar to Swin but involves an extra temporal dimension.

Given an input feature of size $C \times T \times H \times W$ where $C$, $T$, $H$, $W$  respectively represent the channel, time, height, and width dimensions, we first partition it to $\frac{HW}{M^2}$ non-overlapped 3D sub-cubes with the shape of each cube as $T \times M \times M$ (Figure~\ref{fig:partition}(a)-left) and then perform standard multi-head self-attention (MSA) on each sub-cube.
Note that each element of this cube is a $C$-dimensional feature vector, and we omit the channel dimension when describing the partition strategies for simplicity.
Once all the sub-cubes are processed, we merge them back to recover the original shape of the input.
%
%
In order to bridge connections across neighboring cubes, we adopt a shifted-cube partition strategy, which shifts the cubes to top-left by $(\lfloor \frac{M}{2} \rfloor, \lfloor \frac{M}{2} \rfloor)$ pixels (Figure~\ref{fig:partition} (a-right). 

\vspace{1mm}
\noindent\textbf{Separation of space and time.}
Although the above STS can handle video inputs, it may suffer from memory issues when dealing with large cube sizes, \ie, large $T$ or $M$. 
To alleviate this issue, we propose the Sep-STS by separating the spatial-temporal computations into space and time. 
%

First, for the computation in space, given an input feature map with a size of $C \times T \times H \times W$, we first partition it into $\frac{THW}{M^2}$ non-overlapped 2D sub-windows with a size of $M \times M$ as shown in Figure~\ref{fig:partition}(b)-left, and then perform the standard MSA for each sub-window.
For connecting different windows, as we restrict our computations in 2D here, we simply use the shifted window partition strategy of the Swin for each frame as shown in Figure~\ref{fig:partition}(b)-right. 


Second, for the computation in the temporal dimension, we reshape the input feature map into $HW$ temporal vectors with a length of $T$ as shown in Figure~\ref{fig:partition}(c) and perform MSA inside each vector such that the dependencies across frames can be modeled.
This step complements the self-attention in the spatial domain, and thus the two operations need to be used together to process videos.

\vspace{1mm}
\noindent\textbf{Sep-STS block.}
Based on the Sep-STS attention, we devise our main component, the Sep-STS block,
which is composed of separated spatial and temporal attention modules as well as an MLP (Figure~\ref{fig:blocks arch}).
The MLP adopts a two-layer structure and uses the GELU function~\cite{hendrycks2016gaussian} for activation.
Similar to~\cite{liu2021swin}, we apply Layer Normalization (LN)~\cite{ba2016layer} and residual connections~\cite{he2016deep} in this block to stabilize training.
Similar to Swin, the regular and shifted partitions are employed alternatively for consecutive Sep-STS blocks to model long-range spatial-temporal dependencies.

\vspace{1mm}
\noindent\textbf{Memory usage.}
The Sep-STS attention factorizes a computationally expensive operation into two lighter operations in space and time,
which effectively lessens the memory usage reducing from $\mathcal{O}((TMM) \cdot THW)$ of the STS to $\mathcal{O}((T+MM) \cdot THW)$ of our Sep-STS.

During training, compared to the STS baseline, we observe a 26.2\% GPU memory reduction by using our Sep-STS.
As the window size $MM$ is usually much larger than the number of input frames $T$, this reduction ratio essentially relies on $T$ which we set as 4 by default following the settings of the state-of-the-art algorithms~\cite{xu2019quadratic,chi2020all,kalluri2020flavr}.
Since the proposed framework is flexible and can be used for arbitrary number of frames, the Sep-STS can potentially give more significant memory reduction for a larger $T$.
In addition, the space-time separation strategy can also reduce the computational cost similar to the memory usage.
However, as the Sep-STS is naively implemented with two separate PyTorch~\cite{paszke2019pytorch} layers in our experiments, its run-time is in fact similar to that of the STS.
Potentially, optimizing its implementation with a customized CUDA kernel may further improve the efficiency.


\vspace{1mm}
\noindent\textbf{Discussions.}
In this work, we explore the concept of local attention for Transformer-based video interpolation.
Similar concepts have been adopted in other recent methods, such as the local relation network~\cite{hu2019local}, stand-alone network~\cite{ramachandran2019stand}, and Swin~\cite{liu2021swin}.
%
%
Nevertheless, these algorithms are designed for images, and less attention is paid to exploiting local attention mechanisms for videos due to difficulties caused by the extra temporal dimension.
In addition, existing methods mainly focus on image recognition tasks that are generally seen as high-level vision tasks, while in this work we emphasize more on motion modeling and appearance reconstruction.
%
%
%
In this work, we focus on the temporal extension of local attention modules for effective video frame interpolation.
%
We explore the space-time separable local attention, which is in spirit similar to MobileNet~\cite{howard2017mobilenets}  that improves a standard convolution by factorizing it into a depthwise convolution and a pointwise convolution.
Furthermore, we propose a multi-scale kernel-prediction framework to fully exploit the features learned by local attention, as introduced next.

\subsection{Frame Synthesis} \label{sec:frame synthesis}
With the features from the proposed encoder-decoder network, our VFIT synthesizes the output image by predicting spatially-variant kernels to adaptively fuse the source frames.
%
%
Different from existing kernel-based video interpolation methods~\cite{niklaus2017video,niklaus2018video,lee2020adacof,shi2021video}, we propose a multi-scale kernel-prediction framework using the hierarchical feature  $\{F^l, l=0,1,2\}$  as shown in Figure~\ref{fig:FrameWork}.

The frame synthesis network of VFIT is composed of three SynBlocks that make predictions at different scales, and each SynBlock is a kernel prediction network. 
VFIT fuses these multi-scale predictions to generate the final result by:
\begin{align}
	\hat{I}_{0.5}^l &= f_\text{up} (\hat{I}_{0.5}^{l+1})+ O^l,  \\
	O^l &= f^l_\text{syn}(F^l, I^l_{\{-(\lfloor{\frac{T}{2}}\rfloor-1), \cdots, {\lceil{\frac{T}{2}}\rceil}\}}), 
\end{align}
where $l=0, 1, 2$ represent different scales from fine to coarse,
and $f_\text{up}$ denotes the bilinear upsampling function.
The synthesized frame at a finer scale $\hat{I}_{0.5}^l$ can be obtained by merging the upsampled output from the coarse scale ($f_\text{up} (\hat{I}_{0.5}^{l+1})$) and the prediction of the current SynBlock ($O^l$).
The output at the finest scale $\hat{I}_{0.5}^0$ is the final result of our VFIT, \ie, $\hat{I}_{0.5}=\hat{I}_{0.5}^0$, and the initial value $\hat{I}_{0.5}^3=0$.
Here, $f^l_\text{syn}$ is the $l$-th SynBlock which takes the spatial-temporal feature $F^l$ and the frame sequence $I^l_{\{-(\lfloor{\frac{T}{2}}\rfloor-1), \cdots, {\lceil{\frac{T}{2}}\rceil}\}}$ as input, and 
$I^l_t$ represents a frame $I_t$ downsampled by a factor of $2^l$ with bilinear interpolation, where $I^0_t$ is equivalent to the original frame without downsampling.
%


\vspace{1mm}
\noindent\textbf{SynBlock.}
Given the input feature map $F^l \in \mathbb{R}^{C \times T \times H \times W}$, the SynBlock generates its prediction at the $l$-th scale by estimating a set of generalized deformable kernels~\cite{stpan} to aggregate the information from the source frames.

As illustrated in Figure~\ref{fig:FrameWork}, we first unbind $F^l$ at the temporal dimension  to get $T$ separate feature maps for all input frames, denoted as $F^l_{\{-(\lfloor{\frac{T}{2}}\rfloor-1), \cdots, {\lceil{\frac{T}{2}}\rceil}\}}$, and for each frame $t$, $F^l_t \in \mathbb{R}^{C\times H \times W}$.
Then we feed $F^l_t$ into three small 2D CNNs to obtain the per-pixel deformable kernels for frame $I_t^l$, including the kernel weights $W_t^l \in \mathbb{R}^{K\times H \times W}$, horizontal offsets $\alpha_t^l \in \mathbb{R}^{K\times H \times W}$, and vertical offsets $\beta_t^l \in \mathbb{R}^{K\times H \times W}$, where $K$ is the number of sampling locations of each kernel.

With the predicted kernels we obtain the output of the SynBlock at location $(x,y)$ for frame $t$ as:
\begin{align}
	O_t^l(x,y)=
	\sum_{k=1}^{K}W_t^l(k,x,y) I_t^l(x+\alpha_t^l(k,x,y), y+\beta_t^l(k,x,y)), \nonumber
\end{align}
which aggregates neighboring pixels around $(x,y)$ with adaptive weights $W$ similar to~\cite{stpan}.

Finally, we generate the output at scale $l$ by blending $O_t^l$ of all frames  with learned masks.
Specifically, we concatenate the feature maps $F^l_{\{-(\lfloor{\frac{T}{2}}\rfloor-1), \cdots, {\lceil{\frac{T}{2}}\rceil}\}}$ at the channel dimension and send the concatenated features to a small 2D CNN to produce $T$ blending masks $B^l_{\{-(\lfloor{\frac{T}{2}}\rfloor-1), \cdots, {\lceil{\frac{T}{2}}\rceil}\}}$.
Note that we use a softmax function as the last layer of the CNN to normalize the masks along the temporal dimension.
The final output of the SynBlock $f^l_\text{syn}$ is generated by:
\begin{equation}
	O^l  = \sum_t B_t^l \cdot O_t^l.
\end{equation}
Note that this SynBlock can be seen as a multi-frame extension of~\cite{lee2020adacof, shi2021video} that originate from the generalized deformable kernels of STPAN~\cite{xu2021temporal}.

\begin{table*}[t]
	\definecolor{mygray-bg}{gray}{0.9}
	\footnotesize
	\centering
	\renewcommand\tabcolsep{9.0pt}
	\caption{{Quantitative comparisons} on the Vimeo-90K, UCF101, and DAVIS datasets. Numbers in bold indicate the best performance and underscored numbers indicate the second best.}
	\vspace{-1mm}
	\resizebox{\textwidth}{!}{\begin{tabular}{ccccccccc}
			\toprule
			\multicolumn{1}{c}{\multirow{2}*{Method}} 
			&{\multirow{2}*{\# Parameters (M)}}
			&\multicolumn{2}{c}{Vimeo-90K} 
			&\multicolumn{2}{c}{UCF101}
			&\multicolumn{2}{c}{DAVIS} \\ 
			\cmidrule(r){3-4} 
			\cmidrule(r){5-6}
			\cmidrule(r){7-8}
			& & PSNR ($\uparrow$) & SSIM ($\uparrow$) & PSNR ($\uparrow$) & SSIM ($\uparrow$) & PSNR ($\uparrow$) & SSIM ($\uparrow$)\\
			\midrule
			SuperSloMo~\cite{jiang2018super} & 39.6 & 32.90 & 0.957 & 32.33 & 0.960 & 25.65 & 0.857\\
			DAIN~\cite{bao2019depth} & 24.0 & 33.35 & 0.945 & 31.64 & 0.957 & 26.12 & 0.870\\
			SepConv~\cite{niklaus2018video} & 21.6 & 33.60 & 0.944 & 31.97 & 0.943 & 26.21 & 0.857\\
			BMBC~\cite{park2020bmbc} & 11.0 & 34.76 & 0.965 & 32.61 & 0.955 & 26.42 & 0.868\\
			CAIN~\cite{choi2020channel} & 42.8 & 34.83 & 0.970 & 32.52 & 0.968 & 27.21 & 0.873\\
			AdaCoF~\cite{lee2020adacof} & 21.8 & 35.40 & 0.971 & 32.71 & 0.969 & 26.49 & 0.866\\
			QVI~\cite{xu2019quadratic} & 29.2 & 35.15 & 0.971 & 32.89 & {0.970} & 27.17 & 0.874\\
			SoftSplat~\cite{niklaus2020softmax} & \underline{7.7} & 35.76 & 0.972 & 32.89 & {0.970} & 27.42 & {0.878}\\
			FLAVR~\cite{kalluri2020flavr} & 42.4 & 36.30 & 0.975 & 33.33 & \textbf{0.971} & 27.44 & 0.874\\
			\cellcolor{mygray-bg}VFIT-S &\cellcolor{mygray-bg}\textbf{7.5} &\cellcolor{mygray-bg}\underline{36.48} &\cellcolor{mygray-bg}\underline{0.976} &\cellcolor{mygray-bg}\underline{33.36} &\cellcolor{mygray-bg}\textbf{0.971} &\cellcolor{mygray-bg}\underline{27.92} & \cellcolor{mygray-bg}\underline{0.885}\\
			\cellcolor{mygray-bg}VFIT-B &\cellcolor{mygray-bg}29.0 &\cellcolor{mygray-bg}\textbf{36.96} &\cellcolor{mygray-bg}\textbf{0.978} &\cellcolor{mygray-bg}\textbf{33.44} &\cellcolor{mygray-bg}\textbf{0.971} &\cellcolor{mygray-bg}\textbf{28.09} &\cellcolor{mygray-bg}\textbf{0.888}\\
			\bottomrule
	\end{tabular}}
	\vspace{-3mm}
	\label{tab:comparison with SOTA}
\end{table*}

\section{Experiments}
\label{sec:exp}
%



\subsection{Implementation Details}
\label{sec:implement}
\noindent\textbf{Network.} As shown in Figure~\ref{fig:FrameWork}, the VFIT encoder consists of four stages that have 2, 2, 6, and 2 Sep-STS blocks, respectively.
The skip connections between the encoder and decoder are realized with concatenation.
%
For all three SynBlocks, we set the deformable kernel size as $K=5\times5$.
We present two variants of VFIT: a base model VFIT-B and a small one VFIT-S,
where the model size of VFIT-S is about 25\% of VFIT-B.
The two models use the same architecture, and the only difference is the channel dimension of each stage, where we shrink the channels by half for VFIT-S.


\vspace{1mm}
\noindent\textbf{Training.}
For training our network, we employ a simple $\ell_1$ loss: $||I_{0.5}- \hat{I}_{0.5}||$, where $I_{0.5}$ is the ground truth. 
We use the AdaMax optimizer \cite{kingma2014adam} with $\beta_1=0.9, \beta_2=0.999$. The training batch size is set as $4$.
We train the models for $100$ epochs, where the learning rate is initially set as  $2e^{-4}$ and gradually decayed to $1e^{-6}$. 


\begin{figure*}[!t]
	\centering
	\begin{minipage}[h]{0.21\linewidth}
		\centering
		\includegraphics[width=\linewidth]{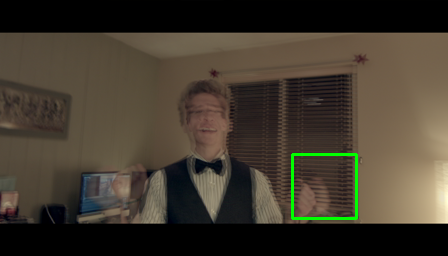}
	\end{minipage}
	\begin{minipage}[h]{0.12\linewidth}
		\centering
		\includegraphics[width=\linewidth]{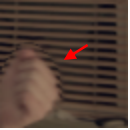}
	\end{minipage}
	\begin{minipage}[h]{0.12\linewidth}
		\centering
		\includegraphics[width=\linewidth]{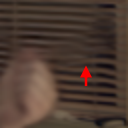}
	\end{minipage}
	\begin{minipage}[h]{0.12\linewidth}
		\centering
		\includegraphics[width=\linewidth]{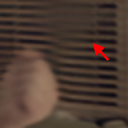}
	\end{minipage}
	\begin{minipage}[h]{0.12\linewidth}
		\centering
		\includegraphics[width=\linewidth]{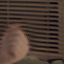}
	\end{minipage}
	\begin{minipage}[h]{0.12\linewidth}
		\centering
		\includegraphics[width=\linewidth]{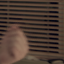}
	\end{minipage}
	\begin{minipage}[h]{0.12\linewidth}
		\centering
		\includegraphics[width=\linewidth]{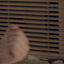}
	\end{minipage}
	\begin{minipage}[h]{0.21\linewidth}
		\centering
		\includegraphics[width=\linewidth]{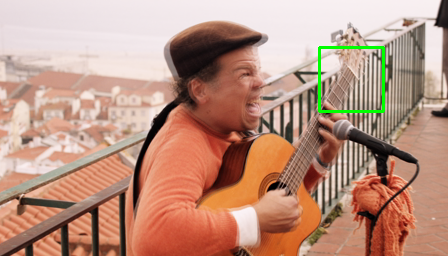}
		\scriptsize{Overlayed}
	\end{minipage}
	\begin{minipage}[h]{0.12\linewidth}
		\centering
		\includegraphics[width=\linewidth]{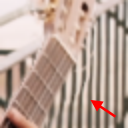}
		\scriptsize{QVI}
	\end{minipage}
	\begin{minipage}[h]{0.12\linewidth}
		\centering
		\includegraphics[width=\linewidth]{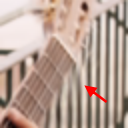}
		\scriptsize{SoftSplat}
	\end{minipage}
	\begin{minipage}[h]{0.12\linewidth}
		\centering
		\includegraphics[width=\linewidth]{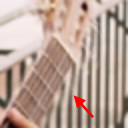}
		\scriptsize{FLAVR}
	\end{minipage}
	\begin{minipage}[h]{0.12\linewidth}
		\centering
		\includegraphics[width=\linewidth]{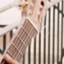}
		\scriptsize{VFIT-S}
	\end{minipage}
	\begin{minipage}[h]{0.12\linewidth}
		\centering
		\includegraphics[width=\linewidth]{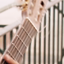}
		\scriptsize{VFIT-B}
	\end{minipage}
	\begin{minipage}[h]{0.12\linewidth}
		\centering
		\includegraphics[width=\linewidth]{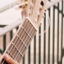}
		\scriptsize{GT}
	\end{minipage}
\vspace{-1mm}
	\caption{{Qualitative comparisons} against the state-of-the-art video interpolation algorithms. The VFIT generates higher-quality results with clearer structures and fewer distortions.}
	\vspace{-2mm}
	\label{fig:qualitative comparisons}
\end{figure*}

\begin{figure*}[!t]
	\centering
	\begin{minipage}[h]{0.21\linewidth}
		\centering
		\includegraphics[width=\linewidth]{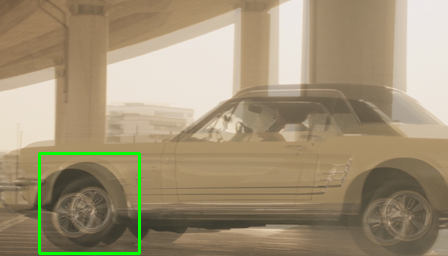}
	\end{minipage}
	\begin{minipage}[h]{0.12\linewidth}
		\centering
		\includegraphics[width=\linewidth]{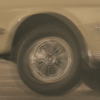}
	\end{minipage}
	\begin{minipage}[h]{0.12\linewidth}
		\centering
		\includegraphics[width=\linewidth]{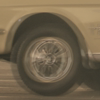}
	\end{minipage}
	\begin{minipage}[h]{0.12\linewidth}
		\centering
		\includegraphics[width=\linewidth]{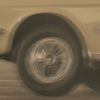}
	\end{minipage}
	\begin{minipage}[h]{0.12\linewidth}
		\centering
		\includegraphics[width=\linewidth]{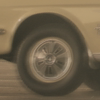}
	\end{minipage}
	\begin{minipage}[h]{0.12\linewidth}
		\centering
		\includegraphics[width=\linewidth]{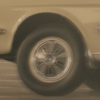}
	\end{minipage}
	\begin{minipage}[h]{0.12\linewidth}
		\centering
		\includegraphics[width=\linewidth]{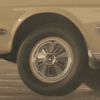}
	\end{minipage}
	\begin{minipage}[h]{0.21\linewidth}
		\centering
		\includegraphics[width=\linewidth]{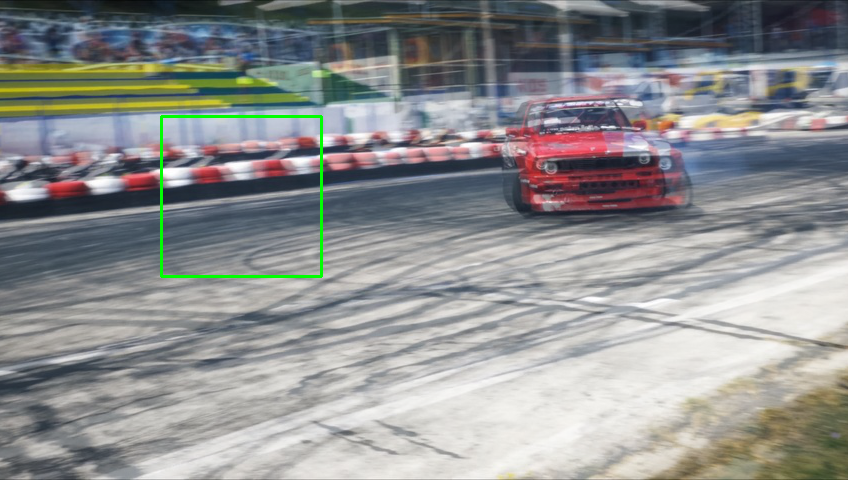}
		\scriptsize{Overlayed}
	\end{minipage}
	\begin{minipage}[h]{0.12\linewidth}
		\centering
		\includegraphics[width=\linewidth]{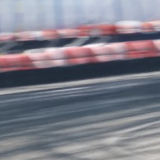}
		\scriptsize{QVI}
	\end{minipage}
	\begin{minipage}[h]{0.12\linewidth}
		\centering
		\includegraphics[width=\linewidth]{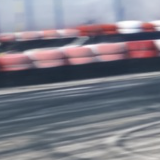}
		\scriptsize{SoftSplat}
	\end{minipage}
	\begin{minipage}[h]{0.12\linewidth}
		\centering
		\includegraphics[width=\linewidth]{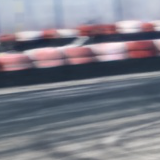}
		\scriptsize{FLAVR}
	\end{minipage}
	\begin{minipage}[h]{0.12\linewidth}
		\centering
		\includegraphics[width=\linewidth]{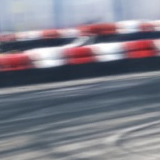}
		\scriptsize{VFIT-S}
	\end{minipage}
	\begin{minipage}[h]{0.12\linewidth}
		\centering
		\includegraphics[width=\linewidth]{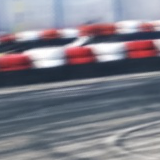}
		\scriptsize{VFIT-B}
	\end{minipage}
	\begin{minipage}[h]{0.12\linewidth}
		\centering
		\includegraphics[width=\linewidth]{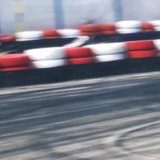}
		\scriptsize{GT}
	\end{minipage}
\vspace{-1mm}
	\caption{{Overlap of interpolated frames and the corresponding ground truth}, where a clearer overlapped image indicates a more accurate prediction. Note that for the second example, as the predictions of the baseline methods and the ground truth are not well aligned, the overlap of the red and white regions presents a blurry pink color. 
	}
	\vspace{-4.5mm}
	\label{fig:qualitative comparisons_motion}
\end{figure*}

\vspace{1mm}
\noindent\textbf{Dataset.}
Similar to~\cite{kalluri2020flavr}, 
we adopt the Vimeo-90K septuplet training set \cite{xue2019video} to learn our models, which consists of $64612$ seven-frame sequences with a resolution of $448 \times 256$. The first, third, fifth, and seventh frames of each sequence correspond to $I_{-1}, I_0, I_1, I_2$ in Figure~\ref{fig:FrameWork} and are used to predict the fourth frame corresponding to $I_{0.5}$. 
For data augmentation, we randomly crop $256 \times 256$ image patches from frames, and perform horizontal and vertical flipping, as well as temporal order reverse.

We evaluate the models on the widely-used benchmark datasets, including the Vimeo-90K septuplet test set~\cite{xue2019video}, UCF101 dataset~\cite{soomro2012ucf101},  and DAVIS dataset~\cite{perazzi2016benchmark}. Following~\cite{xu2019quadratic, kalluri2020flavr}, we report performance on 100 quintuples generated from UCF101 and 2847 quintuples from DAVIS.

%
%
%
%

\subsection{Evaluation against the State of the Arts}
We evaluate the proposed algorithm against the state-of-the-art video interpolation methods: SepConv \cite{niklaus2018video}, DAIN~\cite{bao2019depth}, SuperSloMo \cite{jiang2018super}, CAIN \cite{choi2020channel}, BMBC \cite{park2020bmbc}, AdaCoF \cite{lee2020adacof}, SoftSplat \cite{niklaus2020softmax}, QVI \cite{xu2019quadratic}, and FLAVR \cite{kalluri2020flavr}.
Among these methods, SuperSloMo, DAIN, CAIN, QVI, AdaCoF, and FLAVR are trained on the same training data as our models.
For SepConv and BMBC, as the training code is not available, we directly use the pre-trained models for evaluation. 
The results of SoftSplat~\cite{niklaus2020softmax} are kindly provided by the authors.

We show quantitative evaluations in Table~\ref{tab:comparison with SOTA} where the PSNR and SSIM~\cite{wang2004image} are used for image quality assessment similar to previous works.
%
Thanks to the learning capacity of the Sep-STS block, the proposed VFIT achieves better performance than the evaluated CNN-based methods, demonstrating the superiority of using Transformers for video interpolation.
Specifically, with only 7.5M parameters,  the VFIT-S is able to outperform FLAVR, the best video interpolation method to date, on all the evaluated datasets.
Furthermore, the VFIT-B achieves more significant improvements over FLAVR (0.66 dB on Vimeo-90K and 0.65 dB on DAVIS). 
%
Since the videos of UCF101 have relatively low qualities with low image resolutions and slow motion as explained in~\cite{xu2019quadratic}, our performance gain is less significant. 
Note that the large improvement of the VFIT comes solely from the architecture design without relying on any external information, which differs sharply from several prior works~\cite{bao2019depth,xu2019quadratic,niklaus2020softmax} that use pre-trained optical flow and/or depth models and thus implicitly benefit from additional motion and/or depth labels.

In addition, we provide qualitative comparisons in Figure~\ref{fig:qualitative comparisons},
where the proposed VFIT generates visually more pleasing results with clearer structures and fewer distortions than the baseline approaches.
Moreover, to evaluate the accuracy of the interpolation results, we show overlaps of the interpolated frame and the corresponding ground truth in Figure~\ref{fig:qualitative comparisons_motion}.
The overlapped images of VFIT are much clearer than the baselines, \ie closer to the ground truth, indicating better capabilities of VFIT in motion modeling.



We also present the run-time of our method in Table~\ref{tab:runtime}. 
The run-time performance of the VFIT is on par with the best performing CNN-based algorithms, which facilitates its deployment in vision applications.


\begin{table}[t]
	\footnotesize
	\centering
	\caption{{Run-time of evaluated methods} in seconds per frame. The models are tested on a desktop with an Intel Core i7-8700K CPU and an NVIDIA GTX 2080 Ti GPU. The results are averaged on the Vimeo-90K dataset.}
	\vspace{-1mm}
	\begin{tabular}{cccc|cc}
		\toprule
		Method & BMBC & QVI & FLAVR & VFIT-S & VFIT-B\\
		\midrule
		Run-time & 0.57 & 0.08 & 0.15 & 0.08 & 0.14\\
		\bottomrule
	\end{tabular}
	\vspace{-3mm}
	\label{tab:runtime}
\end{table}

\subsection{Ablation Study}\label{sec:ablation}
We conduct the ablation studies on the Vimeo-90K dataset.
As we notice that the training process converges quickly in the early training stage where differences between models can already be distinguished, we train all models in this study for 20 epochs to accelerate the development and concentrate on the most essential parts of VFIT.


\vspace{1mm}
\noindent\textbf{Local attention.}
In contrast to our model which introduces the local attention mechanism, several recent methods~\cite{chen2021pre,cao2021video} follow the basic structure of conventional Transformers in NLP to use global attention for vision applications,
where the high computational cost of the global attention is circumvented by dividing input into patches and redefining each patch as a new element in self-attention.
In our experiments, we also try this strategy by replacing each Sep-STS block of VFIT-B with a patch-based global-attention block, which is called VFIT-Global.
As shown in Table~\ref{tab:Sep-STS}, the result of VFIT-Global is lower than VFIT-B by as large as 0.84 dB, which emphasizes the essential role of local attention in Transformer-based video frame interpolation.

\vspace{1mm}
\noindent\textbf{Sep-STS.}
To further validate the effectiveness of the Sep-STS block, we compare our VFIT-B with its two variants: 1) VFIT-CNN where all the Sep-STS blocks are replaced by convolutional ResBlocks~\cite{he2016deep}, and each ResBlock is composed of two 3D convolution layers; and 2) VFIT-STS where the Sep-STS block is replaced by its inseparable counterpart, \ie, the STS block.

As shown in Table~\ref{tab:Sep-STS}, while VFIT-CNN uses more than two times  parameters of VFIT-STS, these two models achieve similar results, demonstrating the advantages of using Transformers for video interpolation.
Further, our base model VFIT-B, which uses the proposed Sep-STS as the building block, obtains even better performance than the VFIT-STS.
It should be emphasized that the performance gain is significant as the Sep-STS block is initially designed to reduce memory usage as discussed in Section~\ref{sec:deep feature learning}.
This can be attributed to that the self-attention of the large-size sub-cubes in STS is relatively difficult to learn, and the space-time separation in Sep-STS can serve as a low-rank regularization~\cite{candes2011robust} to remedy this issue.

\begin{figure}[t]
		\centering
		\begin{minipage}[h]{0.30\linewidth}
			\centering
			\includegraphics[width=\linewidth]{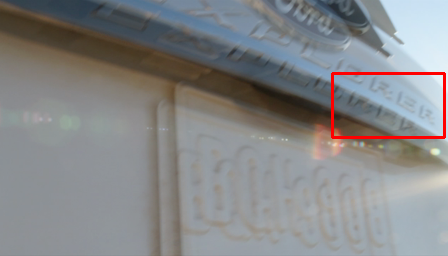}
			\scriptsize{(a) Overlayed}
		\end{minipage}
		\begin{minipage}[h]{0.30\linewidth}
			\centering
			\includegraphics[width=\linewidth]{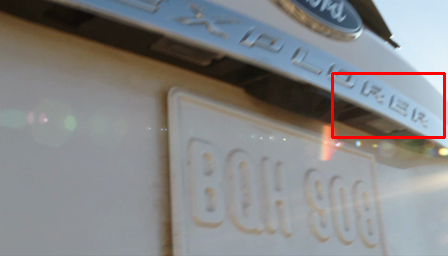}
			\scriptsize{(b) GT}
		\end{minipage}
		\begin{minipage}[h]{0.30\linewidth}
			\centering
			\includegraphics[width=\linewidth]{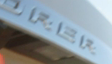}
			\scriptsize{(c) GT-Patch}
		\end{minipage} \\
		\begin{minipage}[h]{0.30\linewidth}
			\centering
			\includegraphics[width=\linewidth]{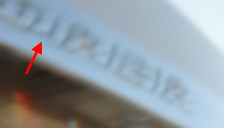}
			\scriptsize{(d) VFIT-CNN}
		\end{minipage}
		\begin{minipage}[h]{0.30\linewidth}
			\centering
			\includegraphics[width=\linewidth]{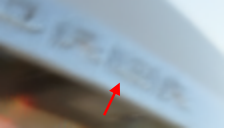}
			\scriptsize{(e) VFIT-STS}
		\end{minipage}
		\begin{minipage}[h]{0.30\linewidth}
			\centering
			\includegraphics[width=\linewidth]{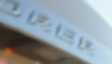}
			\scriptsize{(f) VFIT-B}
		\end{minipage}
		\caption{{Interpolated frames from a video with fast motion.} 
			VFIT-CNN produces severe ghosting artifacts due to it incapability of handling large motion, while the result of VFIT-STS appears blurry. 
			In contrast, the VFIT-B generates a higher-quality intermediate frame closer to the ground truth.}
		\vspace{-2mm}
		\label{fig:fast motions}
	\end{figure}

	
	To better analyze the performance of our models, we further compare with the baselines under different motion conditions. 
	Following~\cite{haris2019recurrent, xiang2020zooming}, we split the Vimeo-90K test set into fast, medium, and slow motions, respectively. 
	Table~\ref{tab:ab_motion} shows VFIT-B outperforms VFIT-CNN by 0.43 dB on fast motion, 0.16 dB on medium motion, and 0.10 dB on slow motion, highlighting the exceptional capability of the proposed Sep-STS in handling challenging large-motion scenarios.
	We also provide interpolated frames from a video with fast motion in Figure~\ref{fig:fast motions} for comparisons.
	
	
	
	
	
	To analyze the effect of different window size of the Sep-STS, we evaluate the VFIT-B with $M=4, 6, 8, 10$, respectively. 
	Table~\ref{tab:Sep-STS} shows, our model performs better as the window size is increased until $M > 8$.
	Thus, we choose $M=8$ as our default setting in this work. 
	


	\begin{table}[t]
		\centering
		\footnotesize
		\caption{\text{Effectiveness of the proposed Sep-STS block.}}
		\vspace{-1mm}
		\begin{tabular}{cccc}
			\toprule
			Method & PSNR \hspace{3mm}&\hspace{3mm}SSIM\hspace{3mm} &\#Parameters (M)\\
			\midrule
			VFIT-B &$\mathbf{36.02}$ &$\mathbf{0.975}$ & $29.0$\\
			VFIT-STS & $35.84$& $0.974$& $29.1$\\
			VFIT-CNN & $35.82$& $0.973$& $65.4$\\
			VFIT-Global &$35.18$ & $0.971$ &$42.4$\\
			\midrule
			$M=4$ &$35.82$ & $0.974$ &$29.0$\\
			$M=6$ &$35.90$ & $0.974$ &$29.0$\\
			$M=8$ &$\mathbf{36.02}$ & $\mathbf{0.975}$ & $29.0$\\
			$M=10$ & $35.93$ & $0.974$ &$29.0$\\
			\bottomrule
		\end{tabular}
		\vspace{-1mm}
		\label{tab:Sep-STS}
	\end{table}
	
	\begin{table}[t]
		\footnotesize
		\centering
		\caption{{Comparison with the base models under different motion conditions.}}
		\vspace{-1mm}
		\resizebox{\columnwidth}{!}{
			\begin{tabular}{cccc}
				\toprule
				Method & Fast &Medium &Slow\\
				\midrule
				VFIT-B &$\mathbf{33.23/0.954}$ & $\mathbf{35.91/0.976}$ &  $\mathbf{38.36/0.987}$\\
				VFIT-STS & $32.91/0.950$ &  $35.77/0.975$ & $38.27/0.987$\\
				VFIT-CNN & $32.80/0.950$ &  $35.75/0.975$ & $38.26/0.987$\\
				VFIT-Global &$32.15/0.945$ &$35.10/0.972$ & $37.62/0.985$\\
				\bottomrule
		\end{tabular}}
		\vspace{-3mm}
		\label{tab:ab_motion}
	\end{table}
	
	\vspace{1mm}
	\noindent\textbf{Multi-scale frame synthesis.}
	 In Section~\ref{sec:frame synthesis}, we propose a multi-scale kernel-prediction network for final frame synthesis.
	To verify the effectiveness of this design, we experiment with a single-scale variant of the VFIT, called VFIT-Single, by removing the second and third SynBlocks in Figure~\ref{fig:FrameWork}.
	This single-scale strategy is essentially similar to the ordinary kernel-prediction networks in~\cite{lee2020adacof,niklaus2017video,niklaus2018video}.
	The PSNR achieved by VFIT-Single is 35.54 dB, which is 0.48 dB lower than our base model VFIT-B. 
	The large performance gap shows the importance of the multi-scale framework for fully realizing the potential of Transformers.
	
	
	Note that we only apply the loss function to the final output, \ie, the finest level output $\hat{I}_{0.5}^0$ of the multi-scale framework as introduced in Section~\ref{sec:implement}.
	Alternatively, one may consider adding supervision to all-scale outputs of the network. 
	However, we empirically find this scheme does not perform well.
	%
	

	\vspace{1mm}
	\noindent\textbf{Resizing modules.}
	As illustrated in Figure~\ref{fig:FrameWork},
	we use 3D convolution and deconvolution layers for downsampling and upsampling the feature maps.
	Motivated by the performance gain of our Sep-STS over CNN-based models, it is of great interest to explore the use of
	Transformer layers as the resizing modules for video frame interpolation.
	
	To answer this question, we adopt the method in~\cite{fan2021multiscale} which introduces a Transformer-based resizing module for video classification by downsampling the query of the self-attention layer.
	To enable Transformer-based upsampling, we extend the idea in~\cite{fan2021multiscale} by upsampling the query with bilinear interpolation.
	%
	%
	We respectively replace the convolution and deconvolution layers of VFIT-B with these Transformer-based downsampling and upsampling modules, and refer to the two variants as VFIT-TD and VFIT-TU.
	As shown in Table~\ref{tab:resize}, both VFIT-TD and VFIT-TU perform slightly worse than our base model with degraded run-time performance, indicating that the current designs of Transformer-based resizing operations in computer vision are less effective for complex motion modeling.
	This is a limitation of our current work, which will be an interesting problem for future research.

	\begin{table}[t]
		\centering
		\footnotesize
		\caption{\text{Comparison with Transformer-based resizing modules.}}
		\begin{tabular}{cccc}
			\toprule
			Method & PSNR & SSIM & Run-time (s)\\
			\midrule
			VFIT-B &$\mathbf{36.02}$ &$\mathbf{0.975}$ &$0.14$\\
			VFIT-TD & $35.92$& $0.974$&$0.17$\\
			VFIT-TU & $35.97$& $0.974$&$0.20$\\
			\bottomrule
		\end{tabular}
		\vspace{-4mm}
		\label{tab:resize}
	\end{table}
	

\section{Conclusion}
In this paper, we propose a parameter, memory, and runtime efficient VFIT framework for video frame interpolation with the state-of-the-art performance. 
	%
	%
A significant part of our effort focuses on extending the local attention mechanism to the spatial-temporal space,  and this module can be integrated in other video processing tasks.
In addition, we demonstrate the effectiveness of a novel space-time separation scheme, which implies the necessity of well-structured regularizations in video Transformers. 
	%
	%
The architecture of VFIT is simple and compact, which can be effectively applied to numerous downstream vision tasks. 

Similar to most existing kernel-based methods~\cite{lee2020adacof, niklaus2017video, niklaus2018video, shi2021video}, we only perform 2$\times$ interpolation with VFIT. However, it can be easily extended to multi-frame interpolation by predicting kernels tied to different time steps or even arbitrary-time interpolation by taking time as an extra input similar to \cite{cheng2021multiple}. 
This will be part of our future work.

\vspace{1mm}
\noindent\textbf{Acknowledgement.}
M.-H. Yang is supported in part by the NSF CAREER Grant \#1149783.

{\small
	\bibliographystyle{ieee_fullname}
	\bibliography{egbib}
}

\end{document}